\documentclass[letterpaper]{article} 
\usepackage{aaai2026}  
\usepackage{times}  
\usepackage{helvet}  
\usepackage{courier}  
\usepackage[hyphens]{url}  
\usepackage{graphicx} 
\usepackage{natbib}  
\usepackage{caption} 
\usepackage{algorithm}
\usepackage{algorithmic}
\usepackage{tabularx}
\usepackage{array}
\usepackage{xcolor} 
\usepackage{colortbl}
\usepackage{amsmath,amssymb}  
\usepackage{pgfplots}
\usepackage{pifont}
\usepackage{booktabs}
\usepackage{newfloat}
\usepackage{listings}
\DeclareCaptionStyle{ruled}{labelfont=normalfont,labelsep=colon,strut=off} 
\floatstyle{ruled}
\newfloat{listing}{tb}{lst}{}
\floatname{listing}{Listing}
\title{SAM2MOT: A Novel Paradigm of Multi-Object Tracking by Segmentation}
\author{
    Junjie Jiang\textsuperscript{\rm 1}\equalcontrib \quad
    Zelin Wang\textsuperscript{\rm 1}\equalcontrib \quad
    Manqi Zhao\textsuperscript{\rm 1} \quad
    Yin Li\textsuperscript{\rm 1} \quad
    DongSheng Jiang\textsuperscript{\rm 1}\thanks{Corresponding author.}
}
\affiliations{
    \textsuperscript{\rm 1}Huawei Cloud\\
    \{jiangjunjie24, wangzelin14, jiangdongsheng1, zhaomanqi, liyin9\}@huawei.com
}

\usepackage{bibentry}

\begin{document}

\maketitle

\begin{abstract}
Inspired by Segment Anything 2, which generalizes segmentation from images to videos, we propose SAM2MOT—a novel segmentation-driven paradigm for multi-object tracking that breaks away from the conventional detection-association framework. In contrast to previous approaches that treat segmentation as auxiliary information, SAM2MOT places it at the heart of the tracking process, systematically tackling challenges like false positives and occlusions. Its effectiveness has been thoroughly validated on major MOT benchmarks. Furthermore, SAM2MOT integrates pre-trained detector, pre-trained segmentor with tracking logic into a zero-shot MOT system that requires no fine-tuning. This significantly reduces dependence on labeled data and paves the way for transitioning MOT research from task-specific solutions to general-purpose systems. Experiments on DanceTrack, UAVDT, and BDD100K show state-of-the-art results. Notably, SAM2MOT outperforms existing methods on DanceTrack by +2.1 HOTA and +4.5 IDF1, highlighting its effectiveness in MOT.
\end{abstract}

\begin{links}
    \link{Code}{https://github.com/TripleJoy/SAM2MOT}
\end{links}

\section{Introduction}

Multi-object tracking (MOT) has long relied on the dominant detection-association paradigm\cite{tbd-sort,tbd-first,tbd-online,tbd-bytetrack,tbd-ocsort}. While this architecture is intuitive, it reveals structural vulnerabilities on two fronts. First, in crowded or occluded scenes, frequent interactions between objects lead to unstable detections, making association highly error-prone. Second, it heavily depends on the detector’s performance—when detections are ambiguous, tracking often collapses. This raises a fundamental question: should we continue to follow a paradigm that has already exposed its inherent limitations?

\definecolor{LightBlue}{RGB}{60, 179, 113}  
\definecolor{LightGreen}{RGB}{30, 144, 255} 
\definecolor{LightOrange}{RGB}{255, 128, 0} 
\definecolor{LightPurple}{RGB}{155, 48, 255} 

\begin{figure}[ht]
    \hspace*{-0.5cm}
    \begin{tikzpicture}
        \begin{axis}[
            xlabel style={font=\ttfamily},  
            ylabel style={font=\ttfamily},  
            ticklabel style={font=\ttfamily},  
            width=9cm, height=7cm,
            xlabel={HOTA},
            ylabel={IDF1},
            title={},
            xmin=42.5, xmax=80,
            ymin=48, ymax=90,
            ylabel style={at={(axis description cs:0.1,0.5)}, anchor=south},
            xlabel style={at={(axis description cs:0.5,0.05)}, anchor=north},
            yticklabel style={font=\scriptsize},
            xticklabel style={font=\scriptsize},
            minor grid style={dashed, black!30},
            minor x tick num=1, 
            minor y tick num=1,
            grid = both,
            legend pos=south east, 
            legend style={font=\ttfamily\tiny},
            legend columns=1,
            legend image post style={scale=0.8}  
        ]
        \addlegendimage{only marks, mark=*, fill=LightBlue, draw=none, opacity=0.6,mark size=3pt}
        \addlegendentry{Fine-tuning for detector}
    
        \addlegendimage{only marks, mark=*, fill=LightGreen, draw=none, opacity=0.6,mark size=3pt}
        \addlegendentry{Training for tracker}
    
        \addlegendimage{only marks, mark=*, fill=LightPurple, draw=none,opacity=0.6, mark size=3pt}
        \addlegendentry{Both}
    
        \addlegendimage{only marks, mark=*, fill=LightOrange, draw=none,opacity=0.6, mark size=3pt}
        \addlegendentry{Neither}
    
        \node[anchor=south] at (axis cs: 50, 82) {\texttt{DanceTrack}};
        
        
        \addplot[
        only marks,
        mark=*,
        mark size=3.14,
        fill=LightBlue,
        opacity=0.6,
        draw=none
        ] coordinates {
            (47.3,52.5) 
        };
        \node at (axis cs:47.3,53.5) [above,font=\ttfamily\scriptsize] {ByteTrack};

        \addplot[
        only marks,
        mark=*,
        mark size=4.02,
        fill=LightBlue,
        opacity=0.6,
        draw=none
        ] coordinates {
            (54.6,54.6) 
        };
    
        \node at (axis cs:54.6,55.7) [above,font=\ttfamily\scriptsize] {OC-SORT};

        \addplot[
            only marks,
            mark=*,
            mark size=3.12,
            fill=LightBlue,
            opacity=0.6,
            draw=none
        ] coordinates {(48.9,50.5)};
        \node at (axis cs:49.8,50.8) [above,font=\ttfamily\scriptsize] {SORT};

        \addplot[
            only marks,
            mark=*,
            mark size=4.02,
            fill=LightGreen,
            opacity=0.6,
            draw=none
        ] coordinates {(54.2,51.5)};
        \node at (axis cs:56.7,50.7) [above,font=\ttfamily\scriptsize] {MOTR};

        \addplot[
            only marks,
            mark=*,
            mark size=5.90,
            fill=LightGreen,
            opacity=0.6,
            draw=none
        ] coordinates {(69.9,71.7)};
        \node at (axis cs:69.3,72.7) [above,font=\ttfamily\scriptsize] {MOTRv2};
        
        \addplot[
            only marks,
            mark=*,
            mark size=6.23,
            fill=LightGreen,
            opacity=0.6,
            draw=none
        ] coordinates {(72.6,74.0)};
        \node at (axis cs:72.6,75.0) [above,font=\ttfamily\scriptsize] {ColTrack};
        
        \addplot[
            only marks,
            mark=*,
            mark size=3.85,
            fill=LightGreen,
            opacity=0.6,
            draw=none
        ] coordinates {(52.7,59.8)};
        \node at (axis cs:52.7,60.8) [above,font=\ttfamily\scriptsize] {FineTrack};
        
        \addplot[
            only marks,
            mark=*,
            mark size=4.72,
            fill=LightGreen,
            opacity=0.6,
            draw=none
        ] coordinates {(62.3,63.0)};
        \node at (axis cs:62.3,64.0) [above,font=\ttfamily\scriptsize] {DiffMOT};
        
        \addplot[
            only marks,
            mark=*,
            mark size=5.43,
            fill=LightPurple,
            opacity=0.6,
            draw=none
        ] coordinates {(66.6,69.7)};
        \node at (axis cs:66.2,70.7) [above,font=\ttfamily\scriptsize] {AED};
        
        \addplot[
            only marks,
            mark=*,
            mark size=6.59,
            fill=LightPurple,
            opacity=0.6,
            draw=none
        ] coordinates {(73.7,79.4)};
        \node at (axis cs:73.3,80.4) [above,font=\ttfamily\scriptsize] {MOTIP};
    
        \addplot[
            only marks,
            mark=*,
            mark size=7.22,
            opacity=0.6,
            fill=LightOrange,
            draw=none
        ] coordinates {(75.8,83.9)};
    
        \node at (axis cs:75.8,85.2) [above,font=\ttfamily\scriptsize] {SAM2MOT};

        \end{axis}
    \end{tikzpicture}
    \caption{IDF1-HOTA-AssA comparisons of different trackers on the test set of DanceTrack, where the horizontal axis represents HOTA, the vertical axis represents IDF1, and the circle radius indicates AssA. This comparison highlights our method's superior capability in associating objects across frames, surpassing all previous trackers.}
\end{figure}

To address the shortcomings of the detection–association paradigm, recent studies propose end-to-end approaches\cite{tbq-transtrack,vistr,tbq-trackformer,tbq-motr} that jointly model detection and association or directly learn identity continuity. Though conceptually promising, these methods face a major limitation—their dependence on large-scale, high-quality tracking data. Unfortunately, since annotating identities across frames is costly and labor-intensive, such data is scarce, leaving end-to-end models prone to unreliable associations and poor generalization under limited supervision.

These challenges compel us to fundamentally rethink the design of multi-object tracking architectures—toward approaches that go beyond detection reliance and generalize effectively in data-scarce scenarios.

Segmentation, though offering pixel-level precision and strong spatial awareness, has long remained peripheral in the MOT community. We argue that the root cause lies in existing methods largely treating segmentation as auxiliary information, lacking systematic modeling and dedicated architectural design. As a result, segmentation-based approaches consistently underperform on major benchmarks, further eroding confidence in their effectiveness. Inspired by Segment Anything 2 (SAM2), we revisit segmentation’s role in video understanding and explore its potential as a core tracking mechanism to overcome current structural limitations.

Building on this insight, we introduce SAM2MOT, a segmentation-driven paradigm for multi-object tracking. It incorporates a cross-object interaction mechanism to systematically model and reduce identity ambiguity under occlusion. Furthermore, we design a unified framework that integrates a pre-trained detector and segmentor with tracking logic, enabling robust zero-shot identity tracking without scene-specific fine-tuning.

To the best of our knowledge, SAM2MOT is the first systematic study of zero-shot tracking in MOT. It introduces a new path toward more generalizable and scalable tracking systems and offers tangible benefits for data construction. By using its zero-shot ability to generate high-quality tracking pre-annotations, manual labeling costs can be greatly reduced, accelerating large-scale dataset creation and promoting progress across the MOT community.

We conduct zero-shot evaluations on the DanceTrack, UAVDT, and BDD100K benchmarks. Without fine-tuning, SAM2MOT achieves state-of-the-art results, surpassing all closed-set fine-tuned methods and showing clear advantages in identity association. These results provide strong evidence for the value of segmentation in MOT and may encourage the community to revisit its role as a core tracking approach.

In conclusion, our contributions are as follows:

\begin{itemize}
    \item \textbf{A Novel Segmentation-Tracking Perspective.} Breaking MOT’s stagnation within detection–association frameworks, we introduce a segmentation-driven tracking paradigm that offers new methodological insights and overcomes data dependence. Unlike previous segmentation-auxiliary approaches, we present the first segmentation-centric framework that systematically tackles major MOT challenges such as large-scale false positives and occlusions.
    \item \textbf{Zero-Shot Tracking Breakthrough.} We architecturally integrate a pre-trained detector and a pre-trained segmentor with tracking logic to achieve a zero-shot MOT system, eliminating scenario-specific fine-tuning with labeled data and marking a disruptive advancement.
    \item \textbf{SOTA Benchmark Results.} SAM2MOT achieves state-of-the-art performance on DanceTrack, UAVDT, and BDD100K. Notably, it reaches 75.8 HOTA and 83.9 IDF1 on DanceTrack, surpassing the second-best method by 2.1 and 4.5, respectively.
\end{itemize}

\begin{figure*}[ht]
    \centering
    \includegraphics[width=1.0\linewidth]{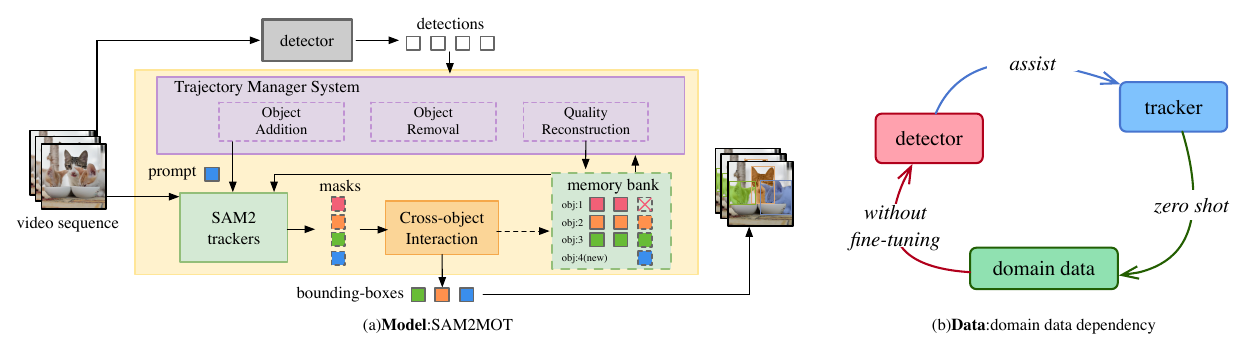}
    \caption{Overview of our \emph{Tracking-by-Segmentation} MOT framework, including: (a) the overall architecture of SAM2MOT; (b) analysis of the relationship between detector, tracker, and data dependency.} 
    \label{fig:overview}
\end{figure*}

\section{Related Works}

\subsection{Conventional MOT Paradigms}

Tracking-by-Detection methods are exemplified by SORT\cite{tbd-sort}, which establishes the paradigm by associating detection boxes across frames using motion modeling and the Hungarian algorithm. Building on this, DeepSORT\cite{tbd-deepsort} integrates ReID features to enhance identity matching robustness. FairMOT\cite{tbd-fairmot} is the first to unify detection and ReID in an end-to-end framework. ByteTrack\cite{tbd-bytetrack} improves robustness in crowded scenes by leveraging low-confidence detections. OC-SORT\cite{tbd-ocsort} proposes an observation-centric Kalman update strategy to better handle occlusions, while StrongSORT\cite{tbd-strongsort} introduces camera motion compensation to improve stability under dynamic viewpoints.

Although these methods have improved in accuracy and robustness, their tight coupling between detection and association makes performance highly dependent on detector quality, demanding extensive scene-specific tuning and hindering generalization.

In contrast, Tracking-by-Query methods use query-based mechanisms to detect new objects and maintain existing tracks within end-to-end transformer architectures. They fall into parallel and sequential categories: VisTR\cite{vistr} and SeqFormer\cite{seqformer} (parallel) process short clips in one pass but suffer from high memory costs, while TransTrack\cite{tbq-transtrack}, TrackFormer\cite{tbq-trackformer}, and MOTR\cite{tbq-motr} (sequential) update queries frame by frame. To enhance temporal modeling, MeMOT\cite{memot} and MeMOTR\cite{memotr} add memory modules or long-term feature injection, and MOTRv3\cite{motrv3} resolves identity conflicts between new and existing objects through improved supervision.

Nevertheless, Tracking-by-Query methods remain limited in real-world scenarios due to their heavy reliance on large-scale, high-quality labeled tracking data, which is often costly and scarce.

\subsection{Segmentation in MOT}

Previous methods that incorporate segmentation mostly treat it an auxiliary information source, typically as a zero-shot alternative to ReID, to enhance object association modules. For example, MASA~\cite{masa} leverages appearance masks extracted by SAM~\cite{sam} to improve the robustness of cross-frame identity matching.

Additionally, several studies explore segmentation-based tracking systems. TrackAnything~\cite{trackanything} combines SAM with XMEM~\cite{xmem} for interactive video object tracking and segmentation, while STAnything~\cite{stanything} further integrates SAM, DeAOT~\cite{deaot}, and Grounding-DINO~\cite{grounding} to build a comprehensive zero-shot tracking framework.

However, we contend that these methods mostly stay at the level of system integration and functional assembly, lacking in-depth modeling or methodological exploration of segmentation as a core tracking mechanism. They often decompose multi-object tracking into independent single-object instances, failing to address challenges like occlusion and false positives, thereby making direct comparison with conventional paradigms difficult. Moreover, with the advent of SAM2~\cite{sam2}, which surpasses earlier combinations such as SAM+XMEM or DeAOT in both structure and performance, the relevance and research value of these approaches have further declined.

\subsection{SAM2 in Tracking}
SAM2~\cite{sam2} extends segmentation from static images to video sequences through a memory bank mechanism, forming an interactive object tracking framework that achieves state-of-the-art performance in single-object video segmentation. Its pipeline starts with a user-provided prompt (points, boxes, or masks) to initialize tracking, extracting appearance features from the key frame and storing them in the memory bank as references. During inference, each frame references the key frame and the previous six frames, performing spatiotemporal matching to predict the object mask and confidence score. The new segmentation is then encoded and added to the memory bank, continually updating and refining the object’s appearance representation.

The memory frame selection strategy in SAM2 is overly simplistic. Subsequent works address this limitation: SAM2-Long\cite{sam2long} employs a decision-tree memory structure for long-range tracking, while SAMURAI\cite{samurai}  improves performance through memory pruning that discards low-confidence entries. Together, these methods continually advance the state of the art in single-object tracking benchmarks.

\section{Method}

\subsection{Overview}

We propose SAM2MOT, focusing on two key challenges: (1) uncovering the role of segmentation in enhancing multi-object tracking, and (2) designing a complete tracking system with zero-shot generalization capability.

Empirically, we find that SAM2, with its pixel-level appearance modeling and implicit motion encoding, provides stronger robustness in single-object tracking than existing methods. However, naïvely extending MOT into multiple parallel single-object trackers introduces major limitations—particularly under occlusion, where degraded appearance quality and missing contextual cues frequently cause identity switches and tracking failures. To address this, we further introduce a Cross-object Interaction module (Section 3.2), which enables communication across objects during parallel tracking and jointly optimizes SAM2’s memory frame selection strategy. This module effectively mitigates occlusion-induced errors and enhances overall tracking accuracy.

Meanwhile, SAM2 inherently operates as an interactive framework relying on user-provided prompts (e.g., boxes, points, or masks). Although a single prompt suffices for single-object tracking, multi-object tracking (MOT) involves the dynamic appearance and disappearance of objects, rendering static initialization inadequate. Therefore, we incorporate a pretrained detector to dynamically generate the required prompts for SAM2. Unlike traditional detection–association paradigms, SAM2 independently generates per-object tracking masks instead of directly inheriting bounding boxes from the detector. This design makes the system less dependent on the precision of a pretrained detector—it merely requires object recognition capability without accurate frame-by-frame localization. Consequently, the SAM2MOT framework achieves zero-shot generalization and decouples tracking performance from specific training data.

Nevertheless, directly connecting a pretrained detector with SAM2 introduces new challenges, including large-scale false positives and degraded long-term tracking quality. To mitigate these issues, we propose the Trajectory Manager System (Section 3.3), which provides a unified mechanism for managing object addition, removal, and quality reconstruction. The overall system architecture is illustrated in Figure \ref{fig:overview}, outlining the integration of the detector, SAM2, and the proposed modules into a unified zero-shot tracking framework.

\subsection{Cross-object Interaction}

Occlusion remains a fundamental challenge in object tracking. When an object B is occluded by another instance object A, SAM2 may erroneously associate the occluded object with the occluder, leading to identity confusion or mismatches. Such identity switches are particularly problematic, as they often propagate over time and prevent object B from being correctly recovered.

The root cause lies in SAM2’s memory bank strategy, which employs a fixed frame selection mechanism: the designated key frame and its six most recent reference frames are used to assist recognition in the current frame. Any error in these frames contaminates the memory bank, subsequently affecting future predictions and progressively compounding errors over time.
In single-object tracking, a typical strategy is to discard low-confidence historical memories to limit error propagation. However, this fixed-threshold approach exhibits clear limitations in multi-object scenarios: a high threshold may prematurely remove valuable information, whereas a low threshold allows erroneous memories to persist. The problem becomes particularly severe during partial occlusions, where tracking errors are more frequent and fixed thresholds fail to distinguish between valid and corrupted memories.

Unlike single-object tracking, SAM2MOT can simultaneously track multiple objects within a frame. Leveraging this capability, we introduce a Cross-object Interaction Module, which enables information exchange between occluding and occluded instances. This mechanism dynamically identifies corrupted tracks during conflicts and removes compromised memory entries in real time, thereby enhancing both tracking accuracy and stability.

Specifically, to detect tracking failures where two objects (e.g., A and B) are mistakenly assigned to the same instance due to occlusion, we measure the overlap between their segmentation masks using the Mask Intersection over Union (mIoU). When the mIoU exceeds 0.8, it indicates substantial overlap, signaling that both trackers are recognizing the same object.

To further determine which object has been incorrectly identified, we design a two-stage analysis method. Assuming object B is misrecognized in the current frame, Stage 1 compares the segmentation confidence scores (i.e., the logit scores defined in SAM2) of objects A and B. If object A’s score is significantly higher (e.g., by more than two points), object B is directly flagged as the incorrectly tracked one. In Stage 2, when confidence scores are similar, the discrepancy usually arises because object A’s confidence declines gradually due to memory degradation, whereas object B’s confidence drops sharply due to sudden occlusion. Based on this observation, we compute the variance of each object’s confidence score over the past N frames (set to 10 in our implementation, see Equation~\ref{eq:var}). A steadily declining score in object A yields a lower variance, allowing object B to be identified as the erroneously tracked instance.

\begin{equation}
\label{eq:var}
    \sigma_{\mathrm{logits}}^2 = \frac{1}{N} \sum_{i=1}^N (\mathrm{logits}_i - \bar{\mathrm{logits}})^2
\end{equation}

\begin{figure}[ht]
    \centering
    \includegraphics[width=1.0\linewidth]{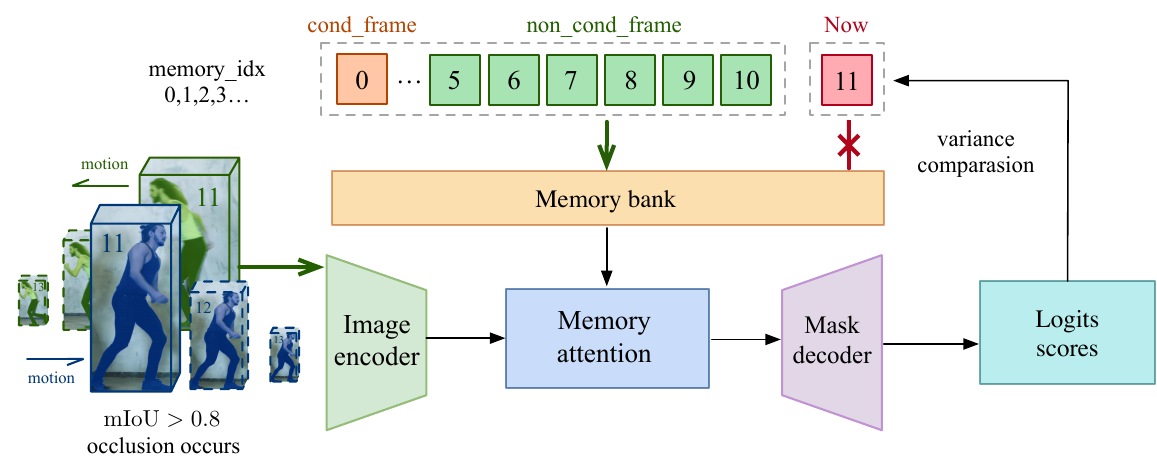}
    \caption{Cross-object interaction pipeline. During motion, when severe occlusion exceeds a predefined threshold(0.8), we identify identity-confused objects by analyzing their logits scores and corresponding variance. The memory entries of such objects in the current frame are then excluded from being written into the memory bank to prevent the propagation of incorrect information.}
    \label{fig:cross-interaction}
\end{figure}

As shown in Figure \ref{fig:cross-interaction}, once object B is identified as having an identity confusion, its current-frame memory information is excluded from memory bank updates to prevent error propagation. Meanwhile, we maintain the filtering mechanism for low-confidence entries but adopt a relatively low threshold, which removes severely degraded appearance features while preserving temporal continuity as much as possible. By combining these two strategies, object B’s memory bank retains only accurate and reliable historical features, enabling robust re-identification and stable tracking even after temporary recognition failures caused by occlusion.

\subsection{Trajectory Manager System}
The motivation for the Trajectory Manager System is presented in Section 3.1. The following subsection details its mechanisms and implementation.

\textbf{Object addition} dynamically generates reliable initial prompts for newly appearing objects, for two main reasons:
(1) Incorrect initialization prompts can cause severe and persistent false tracking in SAM2. Unlike detection–association paradigms, which rely on per-frame detection and can naturally discard false positives if not detected later, SAM2 primarily depends on the initial prompt—once incorrect, the resulting false object will continue to be tracked and is difficult to correct.
(2) Low-quality appearance features (e.g., local features) used for tracking guidance easily cause identity confusion during occlusions or appearance changes.

To address this, SAM2MOT employs a three-stage filtering mechanism built upon the pre-trained detector, enforcing strict criteria to ensure that the generated prompts are both accurate and stable.

Specifically, in each frame, SAM2 generates segmentation masks \(M\) and corresponding tracking boxes for already tracked objects, while the detector outputs a set of detection boxes. The filtering process consists of three steps:

\begin{enumerate}
    \item Discard low-confidence detection boxes using a confidence threshold (e.g., 0.5);
    \item Match the remaining high-confidence boxes with SAM2’s tracking boxes using the Hungarian algorithm, treating unmatched boxes as candidate new objects;
    \item Aggregate and invert all tracked-object masks \(M\) to form $M_{non}$ (Equation~\ref{eq:m_non}) , representing pixels not occupied by tracked objects. For each candidate, compute the overlap area \(p\) between its detection box and $M_{non}$. If the ratio \(p\) /area(box) exceeds a predefined threshold \(r\) (e.g., 0.7), the object is considered sufficiently distinct and initialized as new.
\end{enumerate}

\begin{equation}
\label{eq:m_non}
\mathcal{M}_{\text{non}} = I - \bigcup_{i=1}^{n} \mathcal{M}_i
\end{equation}

\textbf{Object removal} is achieved by setting a frame-based tolerance threshold: if an object remains in a lost state beyond this limit, it is regarded as disappeared and removed. In detection–association paradigms, the lost state is easily determined—if the object is not detected, it is considered lost. However, in SAM2MOT, since SAM2 continuously outputs segmentation results, it may still produce low-confidence masks even when the object has disappeared. Therefore, the key challenge is to accurately determine whether the object is truly lost.

SAM2 outputs a confidence (logit) score for each mask. Based on the score distribution, we define three thresholds—\(\tau_r\), \(\tau_p\) and \(\tau_s\)—to categorize object states into four levels: reliable, pending, suspicious, and lost (Equation~\ref{eq:status}).

\begin{equation}
\label{eq:status}
\begin{aligned}
\mathrm{State} = &\begin{cases}
\mathrm{reliable},   &  \mathrm{logits} > \tau_r,\\
\mathrm{pending},    &  \tau_p < \mathrm{logits} \le \tau_r,\\
\mathrm{suspicious}, &  \tau_s < \mathrm{logits} \le \tau_p,\\
\mathrm{lost},       &  \mathrm{logits} \le \tau_s.
\end{cases}\\[6pt]
\end{aligned}
\end{equation}

A reliable state indicates clear visibility and no occlusion; a pending state suggests mild occlusion or memory degradation, requiring appearance quality reconstruction (described later). A suspicious state usually arises from severe occlusion, where only partial appearance remains—this case is not explicitly handled in this work but remains open for future extensions. The lost state denotes that the object is almost completely occluded or has disappeared, rendering its appearance unavailable.

\textbf{Quality reconstruction} addresses the gradual degradation of prompt information caused by variations in object size, shape, appearance, and position. Since SAM2 references the key frame across subsequent frames, degraded prompts can directly impair tracking. While the original SAM2 allows manual updates of prompt information, SAM2MOT automates this process by leveraging redundant detection cues from the integrated detector.

To minimize computational overhead, reconstruction is triggered only when necessary under two conditions: (1) the object is in the “pending” state; and (2) the current tracking box successfully matches a high-confidence detection box (from the Object Addition stage), ensuring reliable updates. When both conditions are met, the matched high-confidence box updates the object’s keyframe information.

\section{Experiments}
\subsection{Benchmarks}
We evaluate SAM2MOT on three representative benchmark datasets: DanceTrack \cite{data-dancetrack}, UAVDT-MOT \cite{data-uavdt}, and BDD100K-MOT \cite{data-bdd100k}. These datasets collectively encompass a wide spectrum of tracking challenges, including occlusion, diverse motion patterns, and high inter-object similarity. Moreover, they span distinct real-world scenarios—such as indoor dance performances, aerial drone perspectives, and autonomous driving environments—thereby providing a comprehensive evaluation of the model’s adaptability and robustness.

\textbf{Evaluation Metrics.} We adopt HOTA, MOTA, and IDF1 as the primary metrics to comprehensively evaluate the tracking performance of SAM2MOT. For DanceTrack, we further incorporate the sub-metrics DetA and AssA to assess detection and association performance at a finer granularity. Following the official evaluation protocols for BDD100K-MOT and UAVDT-MOT, we report only MOTA and IDF1 to ensure a consistent and fair comparison.

\begin{table*}[t]
    \caption{Comparison with other popular MOT methods on different benchmarks.}
    \label{tab:total_comparison}
    \centering

    \resizebox{1.0\linewidth}{!}{
        \begin{footnotesize}
        \begin{tabular}{l|l|c|c|c|c|c|c|c|c|c|c}
            \toprule
            \multicolumn{1}{l|}{Methods} & Publication  &HOTA$\uparrow$ & MOTA$\uparrow$ & IDF1$\uparrow$ & AssA$\uparrow$ & DetA$\uparrow$  & FN$\downarrow$ & FP$\downarrow$ & IDSW $\downarrow$ & MT$\uparrow$ & ML$\downarrow$ \\  
            
            \midrule
            \midrule
            \underline{\footnotesize \textit{DanceTrack test set}} \\
            
            SORT\cite{tbd-sort} & ICIP2016 & 47.9 & 91.8 & 50.8 & 31.2 & 72.0 
            & - & - & - & - & - \\
            DeepSORT\cite{tbd-deepsort} & ICIP2017 & 45.6 & 87.8 & 47.9 & 29.7 & 71.0 
            & - & - & - & - & - \\
            FairMOT\cite{tbd-fairmot} & IJCV2021 & 39.7 & 82.2 & 40.8 & 23.8 & 66.7 
            & - & - & - & - & - \\
            CenterTrack\cite{centertrack} & ECCV2020 & 41.8 & 86.8 & 35.7 & 22.6 & 78.1 
            & - & - & - & - & - \\
            GTR\cite{gtr} & CVPR2022 & 48.0 & 84.7 & 50.3 & 31.9 & 72.5 
            & - & - & - & - & - \\
            ByteTrack\cite{tbd-bytetrack} & ECCV2022 & 47.3 & 89.5 & 52.5 & 31.4 & 71.6
            & - & - & - & - & - \\
            MOTR\cite{tbq-motr} & ECCV2022 & 54.2 & 79.7 & 51.5 & 40.2 & 73.5 
            & - & - & - & - & - \\
            SUSHI\cite{sushi} & CVPR2023 & 63.3 & 88.7 & 63.4 & 50.1 & 80.1 
            & - & - & - & - & - \\
            MOTRv2\cite{tbq-motrv2} & CVPR2022 & 69.9 & 91.9 & 71.7 & 59.0 & \textbf{83.0} & - & - & - & - & - \\
            ColTrack\cite{coltrack} & ICCV2023 & 72.6 & 92.1 & 74.0 & 62.3 & - 
            & - & - & - & - & - \\
            FineTrack\cite{finetrack} & CVPR2023 & 52.7 & 89.9 & 59.8 & 38.5 & 72.4 
            & - & - & - & - & - \\
            OC-SORT\cite{tbd-ocsort} & CVPR2023 & 54.6 & 89.6 & 54.6 & 40.2 & 80.4 
            & - & - & - & - & - \\
            DiffMOT\cite{diffmot} & CVPR2024 & 62.3 & \textbf{92.8} & 63.0 & 47.2 & 82.5 
            & - & - & - & - & - \\
            Hybrid-SORT\cite{tbd-hybridsort} & AAAI2024 & 65.7 & 91.8 & 67.4 & - & - 
            & - & - & - & - & - \\
            AED\cite{aed} & TIP2025 & 66.6 & 92.2 & 69.7 & 54.3 & 82.0 
            & - & - & - & - & - \\
            MOTIP\cite{motip}& CVPR2025 & 73.7 & 92.7 & 79.4 & 65.9 & 82.6 
            & - & - & - & - & - \\
            \midrule
            SAM2MOT\textit{(Co-DINO-L, no fine-tuning)} & Ours & 75.5 & 89.2 & 83.4 & 71.3 & 80.3 
            & - & - & - & - & - \\
            SAM2MOT\textit{(Grounding-DINO-L, no fine-tuning)} & Ours & \textbf{75.8} & 88.5 & \textbf{83.9} & \textbf{72.2} & 79.7 
            & - & - & - & - & - \\
            
            \midrule
            \midrule
            \underline{\footnotesize \textit{UAVDT-MOT test set}} \\
            
            SiamMOT\cite{siammot} & CVPR2021
            & - & 39.4 & 61.4 & - & -  
            & 176164 & 46903 & 190 & - & - 
            \\
            OC-SORT\cite{tbd-ocsort} & CVPR2023
            & - & 47.5 & 64.9 & - & -
            & 148378 & 47681 & 288 & - & -
            \\
            UAVMOT \cite{uavmot} & CVPR2022  
            & - & 46.4 & 67.3 & - & - 
            & 115940 & 66352  & 456  & 624  & 221 
            \\
            FOLT\cite{folt} & MM2023
            & - & 48.5 & 68.3 & - & -
            & 155696 & \textbf{36429} & 338 & - & -
            \\
            GLOA\cite{gloa} & J-STARS2023
            & - & 49.6 & 68.9 & - & -
            & 115567 & 55822 & 433 & 626 & 220
            \\
            DroneMOT\cite{dronemot} & ICRA2024
            & - & 50.1 & 69.6 & - & -
            & 112548 & 57411 & \textbf{129} & 638 & 178 
            \\
            \midrule
            SAM2MOT\textit{(iou0.5, Co-DINO-L, no fine-tuning)} & Ours 
            & - & 55.6 & 74.4 & - & - 
            & 92504 & 58610 & 141 & 742 & 161 
            \\
            SAM2MOT\textit{(iou0.5, Grounding-DINO-L, no fine-tuning)} & Ours 
            & - & 51.0 & 71.7 & - & - 
            & 103977 & 62906 & 139 & 694 & 189 
            \\
            SAM2MOT\textit{(iou0.4, Co-DINO-L, no fine-tuning)} & Ours 
            & - & \textbf{66.1} & \textbf{79.3} & - & - 
            & \textbf{74586} & 40692 & 136 & \textbf{816} & \textbf{147} 
            \\
            SAM2MOT\textit{(iou0.4, Grounding-DINO-L, no fine-tuning)} & Ours
            & - & 60.9 & 76.6 & - & -
            & 87003 & 45932 & 155 & 767 & 171 
            \\

            \midrule
            \midrule
            \underline{\footnotesize \textit{BDD100K-MOT val set}} \\

            QDtrack\textit{(cls8)} \cite{qdtrack} & TPAMI2023
            & - & 63.5 & 71.5 & - & - 
            & - & - & 6262 & - & - 
            \\
            Unicorn\textit{(cls8)} \cite{unicorn} & ECCV2022
            & -  & 66.6 & 71.3 & - & - 
            & - & - & 10876 & - & - 
            \\
            MOTRv2\textit{(cls8)} \cite{tbq-motrv2} & CVPR2022
            & - & 65.6 & 72.7 & - & - 
            & - & - & - & - & - 
            \\
            UNINEXT\textit{(cls8)} \cite{uniext} & CVPR2023
            & - & \textbf{67.1} & 69.9 & - & - 
            & - & - & 10222 & - & - 
            \\
            MASA\textit{(cls8)} \cite{masa} & CVPR2024
            & - & - & 71.7 & - & - 
            & - & - & - & - & - 
            \\
            \midrule
            SAM2MOT\textit{(cls-8, Co-DINO-L, no fine-tuning)}  & Ours
            & - & 57.5 & 70.8 & - & - 
            & - & - & 5587 & - & - 
            \\
            SAM2MOT\textit{(cls8, Grounding-DINO-L, no fine-tuning)} & Ours
            & - & 44.1 & 63.6 & - & - 
            & - & - & \textbf{5184} & - & - 
            \\
            SAM2MOT\textit{(cls-3, Co-DINO-L, no fine-tuning)} & Ours      
            & - & 63.0 & \textbf{73.7} & - & - 
            & - & - & 5755 & - & - 
            \\
            SAM2MOT\textit{(cls3, Grounding-DINO-L, no fine-tuning)} & Ours
            & - & 58.0 & 71.0 & - & - 
            & - & - & 5464 & - & - 
            \\
            \bottomrule
        \end{tabular}
        \end{footnotesize}
    }
\end{table*}

\subsection{Implementation Details}

We adopt Co-DINO-L(pretrained on COCO)\cite{codino} and Grounding-DINO-L(pretrained on COCO and Object365)\cite{grounding} as the object detectors, and directly applied to all benchmark datasets during testing without additional fine-tuning. Meanwhile, SAM2 utilizes the pretrained SAM2.1-large weights to ensure high-quality segmentation results.

\textbf{Hyperparameter Settings.} A unified configuration is applied across all datasets. In the Cross-object Interaction module, the window size for computing confidence variance is set to \(N = 10\). In the Trajectory Manager System, for Object Addition, the high-confidence threshold is set by default to 0.5 for Co-DINO-L and 0.4 for Grounding-DINO-L, with an overlap ratio threshold of \(r = 0.7\). For Object Removal, the tolerance is fixed at 25 frames, and object state classification is performed based on logit-based thresholds, with \(\tau_r = 8.0\), \(\tau_p = 6.0\), and \(\tau_s = 2.0\).

\begin{table*}[t]
    \caption{Comparison with ByteTrack and OC-SORT using the same detector on different benchmarks.}
    \label{tab:unified_detector_comparison}
    \centering
    \resizebox{1.0\linewidth}{!}{
        \begin{footnotesize}
        \begin{tabular}{l|c|c|c|c|c|c|c|c|c|c|c}
            \toprule
            \multicolumn{1}{l|}{Methods}  &HOTA$\uparrow$ & MOTA$\uparrow$ & IDF1$\uparrow$ & AssA$\uparrow$ & DetA$\uparrow$ & TP$\uparrow$ & FN$\downarrow$ & FP$\downarrow$ & IDSW$\downarrow$ & MT$\uparrow$ & ML$\downarrow$
            \\
            \midrule
            \midrule
            \underline{\footnotesize \textit{DanceTrack test set}} \\
            ByteTrack\textit{(Co-DINO-L)}
            & 56.1 & 87.2 & 56.6 & 40.3 & 78.2
            & 259793 & 29373 & \textbf{6028} & 1650 & 240 & 7
            \\
            ByteTrack\textit{(Grounding-DINO-L)}
            & 53.3 & 86.8 & 53.7 & 37.0 & 77.0
            & 260160 & 29006 & 7281 & 1864 & 233 & 5
            \\
            OC-SORT\textit{(Co-DINO-L)}
            & 56.2 & 86.5 & 56.8 & 40.7 & 77.8
            & - & - & - & - & - & -
            \\
            OC-SORT\textit{(Grounding-DINO-L)}
            & 53.6 & 84.4 & 56.6 & 38.8 & 74.3
            & - & - & - & - & - & -
            \\
            \midrule
            SAM2MOT\textit{(Co-DINO-L)} 
            & 75.5 & \textbf{89.2} & 83.4  & 71.3 & \textbf{80.3}
            & \textbf{274582} & \textbf{14584} & 15653 & \textbf{854} & \textbf{273} & \textbf{2}
            \\
            SAM2MOT\textit{(Grounding-DINO-L)}
            & \textbf{75.8} & 88.5  & \textbf{83.9} & \textbf{72.2} & 79.7  
            & 271472 & 17694 & 14650 & 879 & 264 & 3
            \\
            
            \midrule
            \midrule
            \underline{\footnotesize \textit{UAVDT-MOT test set}} \\

            ByteTrack\textit{(iou0.5, Co-DINO-L)}
            & - & 52.4 & 67.8 & - & - 
            & 204320 & 136586 & 25081 & 452 & 588 & 250 
            \\
            ByteTrack\textit{(iou0.5, Grounding-DINO-L)}  
            & -  & 44.3 & 61.6 & - & -
            & 172820 & 168086 & 21533 & 414 & 465 & 318 
            \\
            ByteTrack\textit{(iou0.4, Co-DINO-L)}   
            & - & 55.4 & 69.1 & - & -
            & 209303 & 131603 & 20098 & 482 & 607 & 240 
            \\
            ByteTrack\textit{(iou0.4, Grounding-DINO-L)} 
            & - & 46.4 & 62.8 & - & -
            & 176507 & 164399 & \textbf{17846} & 477 & 488 & 306 
            \\
            OC-SORT\textit{(iou0.5, Co-DINO-L)}
            & - & 48.5 & 64.5 & - & - 
            & 189117 & 151789 & 23416 & 444 & 534 & 276 
            \\
            OC-SORT\textit{(iou0.5, Grounding-DINO-L)}  
            & - & 39.5 & 56.9 & - & -
            & 154404 & 186502 & 19348 & 368 & 413 & 363 
            \\
            OC-SORT\textit{(iou0.4, Co-DINO-L)}
            & - & 50.8 & 65.6 & - & - 
            & 193014 & 147892 & 19519 & 482 & 555 & 263 
            \\
            OC-SORT\textit{(iou0.4, Grounding-DINO-L)}  
            & - & 41.1 & 57.8 & - & -
            & 157100 & 183806 & 16652 & 417 & 425 & 355 
            \\      
            
            \midrule
            SAM2MOT\textit{(iou0.5, Co-DINO-L)}  
            & - & 55.6 & 74.4 & - & -
            & 248402 & 92504 & 58610 & 141 & 742 & 161 
            \\           
            SAM2MOT\textit{(iou0.5, Grounding-DINO-L)}  
            & - & 51.0 & 71.7 & - & -
            & 236929 & 103977 & 62906 & 139 & 694 & 189 
            \\
            SAM2MOT\textit{(iou0.4, Co-DINO-L)}  
            & - & \textbf{66.1} & \textbf{79.3} & - & -
            & \textbf{266320} & \textbf{74586} & 40692 & \textbf{136} & \textbf{816} & \textbf{147} 
            \\
            SAM2MOT\textit{(iou0.4, Grounding-DINO-L)}  
            & - & 60.9 & 76.6 & -& -
            & 253903 & 87003 & 45932 & 155 & 767 & 171
            \\
            
            \midrule
            \midrule
            \underline{\footnotesize \textit{BDD100K-MOT val set}} \\
            ByteTrack\textit{(cls-3, Co-DINO-L)}
            & - & 54.7 & 60.7 & - & -
            & - & 149852 & \textbf{25710} & 24832  & - & -
            \\ 
            ByteTrack\textit{(cls-3, Grounding-DINO-L)} 
            & -  & 44.8  & 55.6  & - & -
            & - & 179103 & 43437 & 21680 & - & -
            \\
            SAM2MOT\textit{(cls-3, Co-DINO-L)} 
            & - & \textbf{63.0} & \textbf{73.7} & - & - 
            & - & \textbf{125064} & 32965 & 5755 & - & - 
            \\
            SAM2MOT\textit{(cls-3, Grounding-DINO-L)}  
            & - & 58.0 & 71.0 & - & - 
            & - & 138491 & 42070 & \textbf{5464} & - & - 
            \\

            \bottomrule
        \end{tabular}
        \end{footnotesize}
}
\end{table*}

\subsection{State-of-the-Art Comparison}

Table \ref{tab:total_comparison} reports the results of SAM2MOT across different benchmarks, compared with other popular methods.

\textbf{DanceTrack.} The results show that SAM2MOT substantially surpasses existing baselines in handling non-linear object motion, establishing a new state of the art. With Co-DINO-L as the detector, SAM2MOT achieves HOTA = \(75.5\), MOTA = \(89.2\), and IDF1 = \(83.4\). Using Grounding-DINO-L, the corresponding scores are \(75.8\), \(88.5\) and \(83.9\), respectively. Compared to previous best-performing methods, SAM2MOT demonstrates a clear advantage in object association, with HOTA +\(2.1\), IDF1 +\(4.5\), and AssA +\(6.3\).

\textbf{UAVDT-MOT.} On the UAVDT-MOT dataset, using Co-DINO-L as the detector, SAM2MOT achieves MOTA = \(55.6\) and IDF1 = \(74.4\) under the IoU = \(0.5\) evaluation standard. Considering annotation noise in the UAVDT ground truth, we additionally perform an evaluation with a relaxed threshold of IoU = \(0.4\), where MOTA and IDF1 further improve to  \(66.1\) and \(79.4\), respectively. Even under the stricter IoU =  \(0.5\) setting, SAM2MOT attains the best results among all methods. The relaxed IoU = \(0.4\) evaluation better reflects its actual tracking performance, with both metrics showing consistent and significant improvement.

\textbf{BDD100K-MOT.} On the BDD100K-MOT dataset, using Co-DINO-L as the detector, SAM2MOT achieves MOTA = \(57.5\) and IDF1 = \(70.8\) under the eight-class configuration. When evaluated under the three-class setting, the results improve to MOTA = \(63.0\) and IDF1 = \(73.7\). The relatively lower performance in the 8-class setting is attributed to the absence of fine-tuning, which leads the pre-trained detector to confuse visually similar categories. In contrast, under the simplified 3-class scheme, SAM2MOT maintains strong IDF1 performance and significantly reduces ID switches, demonstrating its robust identity association capability.

\subsection{Ablation Study and Analysis}

\textbf{Unified Detector Comparison.} To assess SAM2MOT’s sensitivity to detection accuracy, we compare its performance with two representative detection–association methods—ByteTrack and OC-SORT—using the same detector configuration. Other end-to-end approaches are excluded from this comparison since they do not depend on an independent detector. The experimental setup follows the same configuration described in Section 4.3.

As shown in Table \ref{tab:unified_detector_comparison}, under the same detector and unified settings, SAM2MOT consistently outperforms ByteTrack and OC-SORT across all benchmarks, exhibiting a particularly strong advantage in identity association metrics.

Moreover, the results indicate that, compared with conventional detection–association methods, SAM2MOT can recover additional true positives (TPs) through its internally generated bounding boxes, even when the detector fails—effectively reducing its reliance on precise detection outputs.

\textbf{Component Ablation.} We evaluate the contribution of each module in SAM2MOT on the DanceTrack test set, and the results are summarized in Table \ref{tab:ablation}. The evaluation focuses on three core components: Object Addition (Add), Cross-object Interaction (CoI), and Quality Reconstruction (Q-R). Among them, the Cross-object Interaction module provides the most significant contribution to association accuracy, enabling SAM2MOT to achieve a breakthrough improvement in object correlation.

\begin{table}[t]
    \caption{Ablation study on different strategies, and evaluation metrics on DanceTrack test set.}
    \label{tab:ablation}
    \centering
    \resizebox{1\linewidth}{!}{
        \begin{tabular}{cccccccc}
            \toprule
              \textit{Baseline} & \textit{Add}  & \textit{CoI} & \textit{Q-R} & HOTA & MOTA & IDF1 \\
            \midrule
            \midrule
            \underline{\footnotesize \textit{Co-DINO-L}} \\            
            \ding{51} & \hspace{1em} & \hspace{1em} & \hspace{1em} 
            & 62.9  & 55.6 & 69.6 
            \\
            \ding{51} & \ding{51} & \hspace{1em} & \hspace{1em}
            & 67.9  & 69.7  & 74.4 
            \\
            \ding{51} & \ding{51} & \hspace{1em} & \ding{51} 
            & 69.1 & 69.2 & 76.0 
            \\
            \ding{51} & \ding{51} & \ding{51} & \hspace{1em} 
            & 73.8 & 87.4 & 80.9 
            \\
            \ding{51} & \ding{51} & \ding{51} & \ding{51} 
            & 75.5  & \textbf{89.2}  & 83.4 
            \\
            
            \midrule
            \underline{\footnotesize \textit{Grounding-DINO-L}} \\

            \ding{51} & \hspace{1em} & \hspace{1em} & \hspace{1em} 
            & 60.9 & 50.1 & 66.5
            \\
            \ding{51} & \ding{51} & \hspace{1em} & \hspace{1em} 
            & 67.4 & 67.2 & 74.0
            \\
            \ding{51} & \ding{51} & \hspace{1em} & \ding{51} 
            & 69.0 & 68.7 & 76.2
            \\
            \ding{51} & \ding{51} & \ding{51} & \hspace{1em} 
            & 73.6 & 86.5 & 81.2
            \\
            \ding{51} & \ding{51} & \ding{51} & \ding{51}
            & \textbf{75.8}  & 88.5  & \textbf{83.9}
            \\
            \bottomrule
        \end{tabular}
    }
\end{table}

\section{Limitation.}

\subsection{Inference Efficiency.}
The primary limitation of the current architecture lies in its inference speed. We argue that strategically trading off some runtime efficiency to validate the framework’s potential—similar to pioneering efforts—is both reasonable and worthwhile. Nevertheless, we are actively investigating acceleration strategies along three directions: 

\begin{enumerate}
    \item Replacing the detection backbone with lighter-weight variants (e.g., YOLOE \cite{yoloe}); 
    \item Accelerating the segmentation module by leveraging techniques such as EdgeTAM \cite{edgetam}, which enhances the efficiency of SAM2-based architectures;
    \item Minimizing system-level latency through parallel inference across multiple SAM2 instances.
\end{enumerate}

Furthermore, despite its limited inference efficiency, SAM2MOT demonstrates strong zero-shot generalization capability, effectively mitigating the key bottleneck of video tracking data pre-annotation. This substantially reduces manual labeling effort and provides a practical pathway for constructing large-scale tracking datasets within the community.

\subsection{Holistic Object Modeling.}
SAM2MOT lacks holistic object modeling. When only part of an object is visible (e.g., the head while the body is occluded), it tracks stably but the box covers only the visible region. Detection–association methods, by contrast, infer full bounding boxes as their detectors capture the object’s entire extent. This reflects a core limitation of segmentation-driven frameworks, especially evident on benchmarks like MOTChallenge.

\subsection{Long-term Memory.}
SAM2MOT also struggles with long-term memory. Extended occlusions often cause tracking loss since SAM2, built on short-term memory, encodes only adjacent-frame motion without modeling long-term dependencies. Although SAM2 retains limited long-term capacity, it is unstable. We argue that a more adaptive memory-frame selection could leverage appearance cues for implicit re-identification, improving robustness in long-duration tracking.   

\section{Conclusion}

We propose SAM2MOT, a segmentation-driven paradigm for multi-object tracking that encourages the community to revisit the role of segmentation for tracking. Furthermore, we design a unified tracking framework with strong zero-shot generalization capability, effectively addressing performance degradation under data-scarce conditions and enhancing practical applicability. Extensive experiments on multiple mainstream benchmarks validate the superior identity association capability of SAM2MOT.   

\clearpage
\bibliography{aaai2026}

@inproceedings{tbd-first,
  title={Tracking without bells and whistles},
  author={Bergmann, Philipp and others},
  booktitle={Proceedings of the IEEE/CVF international conference on computer vision},
  pages={941--951},
  year={2019}
}

@inproceedings{tbd-sort,
  title={Simple online and realtime tracking},
  author={Bewley, Alex and Ge, Zongyuan and Ott, Lionel and Ramos, Fabio and Upcroft, Ben},
  booktitle={2016 IEEE international conference on image processing (ICIP)},
  pages={3464--3468},
  year={2016},
  organization={Ieee}
}

@inproceedings{tbd-online,
  title={Track to detect and segment: An online multi-object tracker},
  author={Wu, Jialian and Cao, Jiale and Song, Liangchen and Wang, Yu and Yang, Ming and Yuan, Junsong},
  booktitle={Proceedings of the IEEE/CVF conference on computer vision and pattern recognition},
  pages={12352--12361},
  year={2021}
}

@inproceedings{tbd-hybridsort,
  title={Hybrid-sort: Weak cues matter for online multi-object tracking},
  author={Yang, Mingzhan and Han, Guangxin and Yan, Bin and Zhang, Wenhua and Qi, Jinqing and Lu, Huchuan and Wang, Dong},
  booktitle={Proceedings of the AAAI conference on artificial intelligence},
  volume={38},
  pages={6504--6512},
  year={2024}
}

@inproceedings{tbd-bytetrack,
  title={Bytetrack: Multi-object tracking by associating every detection box},
  author={Zhang, Yifu and Sun, Peize and Jiang, Yi and Yu, Dongdong and Weng, Fucheng and Yuan, Zehuan and Luo, Ping and Liu, Wenyu and Wang, Xinggang},
  booktitle={European conference on computer vision},
  pages={1--21},
  year={2022},
  organization={Springer}
}

@inproceedings{tbd-ocsort,
  title={Observation-centric sort: Rethinking sort for robust multi-object tracking},
  author={Cao, Jinkun and Pang, Jiangmiao and Weng, Xinshuo and Khirodkar, Rawal and Kitani, Kris},
  booktitle={Proceedings of the IEEE/CVF conference on computer vision and pattern recognition},
  pages={9686--9696},
  year={2023}
}

@article{tbd-fairmot,
  title={Fairmot: On the fairness of detection and re-identification in multiple object tracking},
  author={Zhang, Yifu and Wang, Chunyu and Wang, Xinggang and Zeng, Wenjun and Liu, Wenyu},
  journal={International journal of computer vision},
  volume={129},
  pages={3069--3087},
  year={2021},
  publisher={Springer}
}

@article{tbd-strongsort,
  title={Strongsort: Make deepsort great again},
  author={Du, Yunhao and Zhao, Zhicheng and Song, Yang and Zhao, Yanyun and Su, Fei and Gong, Tao and Meng, Hongying},
  journal={IEEE Transactions on Multimedia},
  volume={25},
  pages={8725--8737},
  year={2023},
  publisher={IEEE}
}

@article{tbd-deepsort,
  title={DeepSort: deep convolutional networks for sorting haploid maize seeds},
  author={Veeramani, Balaji and Raymond, John W and Chanda, Pritam},
  journal={BMC bioinformatics},
  volume={19},
  pages={1--9},
  year={2018},
  publisher={Springer}
}

@inproceedings{tbq-trackformer,
  title={Trackformer: Multi-object tracking with transformers},
  author={Meinhardt, Tim and Kirillov, Alexander and Leal-Taixe, Laura and Feichtenhofer, Christoph},
  booktitle={Proceedings of the IEEE/CVF conference on computer vision and pattern recognition},
  pages={8844--8854},
  year={2022}
}

@article{tbq-transtrack,
  title={Transtrack: Multiple object tracking with transformer},
  author={Sun, Peize and Cao, Jinkun and Jiang, Yi and Zhang, Rufeng and Xie, Enze and Yuan, Zehuan and Wang, Changhu and Luo, Ping},
  journal={arXiv preprint arXiv:2012.15460},
  year={2020}
}

@inproceedings{tbq-motr,
  title={Motr: End-to-end multiple-object tracking with transformer},
  author={Zeng, Fangao and Dong, Bin and Zhang, Yuang and Wang, Tiancai and Zhang, Xiangyu and Wei, Yichen},
  booktitle={European conference on computer vision},
  pages={659--675},
  year={2022},
  organization={Springer}
}

@inproceedings{tbq-motrv2,
  title={Motrv2: Bootstrapping end-to-end multi-object tracking by pretrained object detectors},
  author={Zhang, Yuang and Wang, Tiancai and Zhang, Xiangyu},
  booktitle={Proceedings of the IEEE/CVF conference on computer vision and pattern recognition},
  pages={22056--22065},
  year={2023}
}

@inproceedings{sam,
  title={Segment anything},
  author={Kirillov, Alexander and Mintun, Eric and Ravi, Nikhila and Mao, Hanzi and Rolland, Chloe and Gustafson, Laura and Xiao, Tete and Whitehead, Spencer and Berg, Alexander C and Lo, Wan-Yen and others},
  booktitle={Proceedings of the IEEE/CVF international conference on computer vision},
  pages={4015--4026},
  year={2023}
}

@article{sam2,
  title={Sam 2: Segment anything in images and videos},
  author={Ravi, Nikhila and Gabeur, Valentin and Hu, Yuan-Ting and Hu, Ronghang and Ryali, Chaitanya and Ma, Tengyu and Khedr, Haitham and R{\"a}dle, Roman and Rolland, Chloe and Gustafson, Laura and others},
  journal={arXiv preprint arXiv:2408.00714},
  year={2024}
}

@article{samurai,
  title={Samurai: Adapting segment anything model for zero-shot visual tracking with motion-aware memory},
  author={Yang, Cheng-Yen and Huang, Hsiang-Wei and Chai, Wenhao and Jiang, Zhongyu and Hwang, Jenq-Neng},
  journal={arXiv preprint arXiv:2411.11922},
  year={2024}
}

@article{sam2long,
  title={Sam2long: Enhancing sam 2 for long video segmentation with a training-free memory tree},
  author={Ding, Shuangrui and Qian, Rui and Dong, Xiaoyi and Zhang, Pan and Zang, Yuhang and Cao, Yuhang and Guo, Yuwei and Lin, Dahua and Wang, Jiaqi},
  journal={arXiv preprint arXiv:2410.16268},
  year={2024}
}

@inproceedings{data-dancetrack,
  title={Dancetrack: Multi-object tracking in uniform appearance and diverse motion},
  author={Sun, Peize and Cao, Jinkun and Jiang, Yi and Yuan, Zehuan and Bai, Song and Kitani, Kris and Luo, Ping},
  booktitle={Proceedings of the IEEE/CVF conference on computer vision and pattern recognition},
  pages={20993--21002},
  year={2022}
}

@inproceedings{data-bdd100k,
  title={Bdd100k: A diverse driving dataset for heterogeneous multitask learning},
  author={Yu, Fisher and Chen, Haofeng and Wang, Xin and Xian, Wenqi and Chen, Yingying and Liu, Fangchen and Madhavan, Vashisht and Darrell, Trevor},
  booktitle={Proceedings of the IEEE/CVF conference on computer vision and pattern recognition},
  pages={2636--2645},
  year={2020}
}

@inproceedings{data-uavdt,
  title={The unmanned aerial vehicle benchmark: Object detection and tracking},
  author={Du, Dawei and Qi, Yuankai and Yu, Hongyang and Yang, Yifan and Duan, Kaiwen and Li, Guorong and Zhang, Weigang and Huang, Qingming and Tian, Qi},
  booktitle={Proceedings of the European conference on computer vision (ECCV)},
  pages={370--386},
  year={2018}
}

@inproceedings{vistr,
  title={End-to-end video instance segmentation with transformers},
  author={Wang, Yuqing and Xu, Zhaoliang and Wang, Xinlong and Shen, Chunhua and Cheng, Baoshan and Shen, Hao and Xia, Huaxia},
  booktitle={Proceedings of the IEEE/CVF conference on computer vision and pattern recognition},
  pages={8741--8750},
  year={2021}
}

@inproceedings{seqformer,
  title={Seqformer: Sequential transformer for video instance segmentation},
  author={Wu, Junfeng and Jiang, Yi and Bai, Song and Zhang, Wenqing and Bai, Xiang},
  booktitle={European Conference on Computer Vision},
  pages={553--569},
  year={2022},
  organization={Springer}
}

@inproceedings{memot,
  title={Memot: Multi-object tracking with memory},
  author={Cai, Jiarui and Xu, Mingze and Li, Wei and Xiong, Yuanjun and Xia, Wei and Tu, Zhuowen and Soatto, Stefano},
  booktitle={Proceedings of the IEEE/CVF conference on computer vision and pattern recognition},
  pages={8090--8100},
  year={2022}
}

@inproceedings{memotr,
  title={MeMOTR: Long-term memory-augmented transformer for multi-object tracking},
  author={Gao, Ruopeng and Wang, Limin},
  booktitle={Proceedings of the IEEE/CVF International Conference on Computer Vision},
  pages={9901--9910},
  year={2023}
}

@article{motrv3,
  title={Motrv3: Release-fetch supervision for end-to-end multi-object tracking},
  author={Yu, En and Wang, Tiancai and Li, Zhuoling and Zhang, Yuang and Zhang, Xiangyu and Tao, Wenbing},
  journal={arXiv preprint arXiv:2305.14298},
  year={2023}
}

@inproceedings{centertrack,
  title={Tracking objects as points},
  author={Zhou, Xingyi and Koltun, Vladlen and Kr{\"a}henb{\"u}hl, Philipp},
  booktitle={European conference on computer vision},
  pages={474--490},
  year={2020},
  organization={Springer}
}

@article{qdtrack,
  title={Qdtrack: Quasi-dense similarity learning for appearance-only multiple object tracking},
  author={Fischer, Tobias and Huang, Thomas E and Pang, Jiangmiao and Qiu, Linlu and Chen, Haofeng and Darrell, Trevor and Yu, Fisher},
  journal={IEEE Transactions on Pattern Analysis and Machine Intelligence},
  volume={45},
  number={12},
  pages={15380--15393},
  year={2023},
  publisher={IEEE}
}

@inproceedings{gtr,
  title={Global tracking transformers},
  author={Zhou, Xingyi and Yin, Tianwei and Koltun, Vladlen and Kr{\"a}henb{\"u}hl, Philipp},
  booktitle={Proceedings of the IEEE/CVF conference on computer vision and pattern recognition},
  pages={8771--8780},
  year={2022}
}

@article{aed,
  title={Associate everything detected: Facilitating tracking-by-detection to the unknown},
  author={Fang, Zimeng and Liang, Chao and Zhou, Xue and Zhu, Shuyuan and Li, Xi},
  journal={IEEE Transactions on Image Processing},
  year={2025},
  publisher={IEEE}
}

@inproceedings{coltrack,
  title={Collaborative tracking learning for frame-rate-insensitive multi-object tracking},
  author={Liu, Yiheng and Wu, Junta and Fu, Yi},
  booktitle={Proceedings of the IEEE/CVF international conference on computer vision},
  pages={9964--9973},
  year={2023}
}

@inproceedings{motip,
  title={Multiple object tracking as id prediction},
  author={Gao, Ruopeng and Qi, Ji and Wang, Limin},
  booktitle={Proceedings of the Computer Vision and Pattern Recognition Conference},
  pages={27883--27893},
  year={2025}
}

@inproceedings{diffmot,
  title={Diffmot: A real-time diffusion-based multiple object tracker with non-linear prediction},
  author={Lv, Weiyi and Huang, Yuhang and Zhang, Ning and Lin, Ruei-Sung and Han, Mei and Zeng, Dan},
  booktitle={Proceedings of the IEEE/CVF Conference on Computer Vision and Pattern Recognition},
  pages={19321--19330},
  year={2024}
}

@inproceedings{sushi,
  title={Unifying short and long-term tracking with graph hierarchies},
  author={Cetintas, Orcun and Bras{\'o}, Guillem and Leal-Taix{\'e}, Laura},
  booktitle={Proceedings of the IEEE/CVF conference on computer vision and pattern recognition},
  pages={22877--22887},
  year={2023}
}

@inproceedings{finetrack,
  title={Focus on details: Online multi-object tracking with diverse fine-grained representation},
  author={Ren, Hao and Han, Shoudong and Ding, Huilin and Zhang, Ziwen and Wang, Hongwei and Wang, Faquan},
  booktitle={Proceedings of the IEEE/CVF conference on computer vision and pattern recognition},
  pages={11289--11298},
  year={2023}
}

@inproceedings{masa,
  title={Matching anything by segmenting anything},
  author={Li, Siyuan and Ke, Lei and Danelljan, Martin and Piccinelli, Luigi and Segu, Mattia and Van Gool, Luc and Yu, Fisher},
  booktitle={Proceedings of the IEEE/CVF Conference on Computer Vision and Pattern Recognition},
  pages={18963--18973},
  year={2024}
}

@inproceedings{unicorn,
  title={Towards grand unification of object tracking},
  author={Yan, Bin and Jiang, Yi and Sun, Peize and Wang, Dong and Yuan, Zehuan and Luo, Ping and Lu, Huchuan},
  booktitle={European conference on computer vision},
  pages={733--751},
  year={2022},
  organization={Springer}
}

@inproceedings{uniext,
  title={Universal instance perception as object discovery and retrieval},
  author={Yan, Bin and Jiang, Yi and Wu, Jiannan and Wang, Dong and Luo, Ping and Yuan, Zehuan and Lu, Huchuan},
  booktitle={Proceedings of the IEEE/CVF Conference on Computer Vision and Pattern Recognition},
  pages={15325--15336},
  year={2023}
}

@inproceedings{siammot,
  title={Siammot: Siamese multi-object tracking},
  author={Shuai, Bing and Berneshawi, Andrew and Li, Xinyu and Modolo, Davide and Tighe, Joseph},
  booktitle={Proceedings of the IEEE/CVF conference on computer vision and pattern recognition},
  pages={12372--12382},
  year={2021}
}

@inproceedings{uavmot,
  title={Multi-object tracking meets moving UAV},
  author={Liu, Shuai and Li, Xin and Lu, Huchuan and He, You},
  booktitle={Proceedings of the IEEE/CVF Conference on Computer Vision and Pattern Recognition},
  pages={8876--8885},
  year={2022}
}

@inproceedings{folt,
  title={Folt: Fast multiple object tracking from uav-captured videos based on optical flow},
  author={Yao, Mufeng and Wang, Jiaqi and Peng, Jinlong and Chi, Mingmin and Liu, Chao},
  booktitle={Proceedings of the 31st ACM International Conference on Multimedia},
  pages={3375--3383},
  year={2023}
}

@article{gloa,
  title={Global-local and occlusion awareness network for object tracking in UAVs},
  author={Shi, Lukui and Zhang, Qingrui and Pan, Bin and Zhang, Jun and Su, Yuanchao},
  journal={IEEE Journal of Selected Topics in Applied Earth Observations and Remote Sensing},
  volume={16},
  pages={8834--8844},
  year={2023},
  publisher={IEEE}
}

@inproceedings{dronemot,
  title={Dronemot: Drone-based multi-object tracking considering detection difficulties and simultaneous moving of drones and objects},
  author={Wang, Peng and Wang, Yongcai and Li, Deying},
  booktitle={2024 IEEE International Conference on Robotics and Automation (ICRA)},
  pages={7397--7404},
  year={2024},
  organization={IEEE}
}

@inproceedings{xmem,
  title={{XMem}: Long-Term Video Object Segmentation with an Atkinson-Shiffrin Memory Model},
  author={Cheng, Ho Kei and Alexander G. Schwing},
  booktitle={ECCV},
  year={2022}
}

@inproceedings{deaot,
  title={Decoupling Features in Hierarchical Propagation for Video Object Segmentation},
  author={Yang, Zongxin and Yang, Yi},
  booktitle={Advances in Neural Information Processing Systems (NeurIPS)},
  year={2022}
}

@article{grounding,
  title={Grounding dino: Marrying dino with grounded pre-training for open-set object detection},
  author={Liu, Shilong and Zeng, Zhaoyang and Ren, Tianhe and Li, Feng and Zhang, Hao and Yang, Jie and Li, Chunyuan and Yang, Jianwei and Su, Hang and Zhu, Jun and others},
  journal={arXiv preprint arXiv:2303.05499},
  year={2023}
}

@misc{trackanything,
      title={Track Anything: Segment Anything Meets Videos}, 
      author={Jinyu Yang and Mingqi Gao and Zhe Li and Shang Gao and Fangjing Wang and Feng Zheng},
      year={2023},
      eprint={2304.11968},
      archivePrefix={arXiv},
      primaryClass={cs.CV}
}

@article{stanything,
  title={Segment and Track Anything},
  author={Cheng, Yangming and Li, Liulei and Xu, Yuanyou and Li, Xiaodi and Yang, Zongxin and Wang, Wenguan and Yang, Yi},
  journal={arXiv preprint arXiv:2305.06558},
  year={2023}
}

@inproceedings{codino,
  title={Detrs with collaborative hybrid assignments training},
  author={Zong, Zhuofan and Song, Guanglu and Liu, Yu},
  booktitle={Proceedings of the IEEE/CVF international conference on computer vision},
  pages={6748--6758},
  year={2023}
}

@article{edgetam,
  title={EdgeTAM: On-Device Track Anything Model},
  author={Zhou, Chong and Zhu, Chenchen and Xiong, Yunyang and Suri, Saksham and Xiao, Fanyi and Wu, Lemeng and Krishnamoorthi, Raghuraman and Dai, Bo and Loy, Chen Change and Chandra, Vikas and Soran, Bilge},
  journal={arXiv preprint arXiv:2501.07256},
  year={2025}
}

@article{yoloe,
  title={Yoloe: Real-time seeing anything},
  author={Wang, Ao and Liu, Lihao and Chen, Hui and Lin, Zijia and Han, Jungong and Ding, Guiguang},
  journal={arXiv preprint arXiv:2503.07465},
  year={2025}
}

\begin{figure*}[ht]
    \centering
    \includegraphics[width=0.98\linewidth]{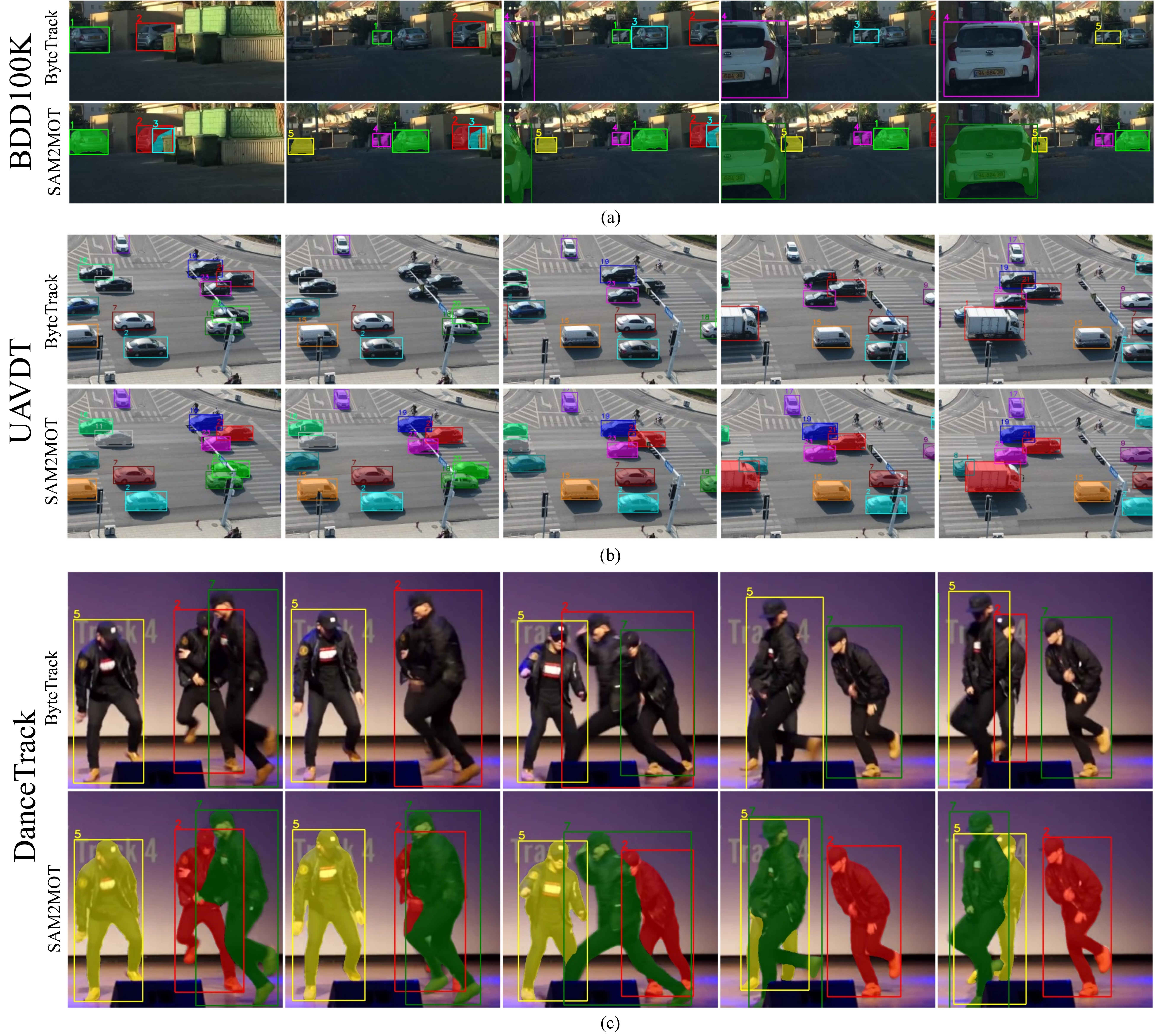}
    \caption{Sample tracking results visualization of ByteTrack and SAM2MOT using the same detector on DanceTrack, BDD100K-MOT and UAVDT-MOT. The results indicate that SAM2MOT significantly outperforms ByteTrack in association performance under scenarios involving camera motion, detector degradation, and occlusion.} 
    \label{fig:visual}
\end{figure*}   
\end{document}